\documentstyle[darpasls,times]{article}

\newcommand {\sub}    [1] {$_{#1}$}       %subscript
\newcommand {\fsse}[2] {\fbox{#1\sub{#2}}}

\newcommand {\vsp}{\vspace*{.07in}}
\newcommand {\kitem}[1]
  {\hspace*{.35in}  $-$ \parbox[t]{2.7in}{#1}}

\newcommand {\jitem}[1]
  {\hspace*{.15in} $\bullet$ \parbox[t]{3in}{#1}}

\newcommand {\jjitem}[1]
  {\vsp\newline\hspace*{.15in} $\bullet$ \parbox[t]{3in}{#1}}

\newcommand {\jenum}[2]
  {\hspace*{.15in} #1 \parbox[t]{3in}{#2}}

\newcommand {\one}[0] {$\ $}
\newcommand {\two}[0] {$\ \ $}
\newcommand {\thr}[0] {$\ \ \ $}

\newcommand {\ul}[1] {\underline{#1}}

\begin{document}

\title{An investigation into the correlation of cue phrases,
unfilled pauses and the structuring of spoken discourse}

\author{Janet Cahn}

\institution{Media Laboratory \\
Massachusetts Institute of Technology \\
Cambridge, MA 02139 \\
{\small \it cahn@media.mit.edu}}

\maketitle

\begin{abstract}
Expectations about the correlation of cue phrases, the duration of
unfilled pauses and the structuring of spoken discourse are framed in
light of Grosz and Sidner's theory of discourse and are tested for a
directions-giving dialogue.  The results suggest that cue phrase and
discourse structuring tasks may align, and show a correlation for
pause length and some of the modifications that speakers can make to
discourse structure.
\end{abstract}

\section{Introduction}
Because an utterance is best understood in the context in which it is
delivered, its interpreters must be able to identify the relevant
context and recognize when it is altered, supplanted or revived. The
transient nature of speech makes this task difficult. However, the
difficulty is alleviated by the abundance of lexical and prosodic cues
available to a speaker for communicating the location and type of
contextual change.  The investigation of the interaction between these
cues presupposes a theory of contextual change in discourse.  The
theory relating attention, intentions and discourse
structure\cite{GROSZ-SIDNER} is particularly useful because it
provides a computational account of the current context and the
mechanisms of contextual change.  This account frames the questions I
investigate about the correlation between between lexical and prosodic
cues.  In particular, the theory motivates the selection of the {\it
cue phrase}\cite{GROSZ-SIDNER} --- a word or phrase whose relevance is
to structural or rhetorical relations, rather than topic --- and the
{\it unfilled pause} (silent pause) as significant indicators of
discourse structure.

\section{The tripartite nature of discourse}

To explain the organization of a discourse into topics and subtopics,
Grosz and Sidner postulate three interrelated components of discourse
--- a linguistic structure, an intentional structure and an
attentional state\cite{GROSZ-SIDNER}.  In the {\it linguistic
structure}, the linear sequence of utterances becomes hierarchical ---
utterances aggregate into discourse segments, and the discourse
segments are organized hierarchically according to the relations among
the purposes or {\it discourse intentions}\footnote{Discourse
intentions are those goals or intentions intended to be recognized by
each participant as the purpose to which the current segment of talk
is devoted.} that each satisfies.

The relations among discourse intentions are captured in the {\it
intentional structure}.  It is this organization that is mirrored by
the linguistic structure of utterances.  However, while the linguistic
structure organizes the verbatim content of discourse segments, the
intentional structure contains only the intentions that underlie each
segment.  The supposition of an intentional structure explains how
discourse coherence is preserved in the absence of a complete history
of the discourse.  Rather, discourse participants summarize the
verbatim contents of a discourse segment by the discourse intention it
satisfies.  The contents of a discourse segment are collapsed into an
intention, and intentions themselves may be collapsed into intentions
of larger scope.

The discourse intention of greatest scope is the Discourse Purpose
(DP), the reason for initiating a discourse.  Within this, discourse
segments are introduced to fulfill a particular Discourse Segment
Purpose (DSP) and thereby contribute to the satisfaction of the
overall DP.  A segment terminates when its DSP is satisfied. 
Similarly, a discourse terminates when the DP that initiated it is
satisfied.

The {\it attentional state} is the third component of the tripartite
theory. It models the foci of attention that exist during the
construction of intentional structures.  The global focus of attention
encompasses those entities relevant to the discourse segment currently
under construction, while the local focus (also called the {\it
center}\cite{CENTERING}) is the currently most salient entity in the
discourse segment.  The local focus may change from utterance to
utterance, while the global focus (i.e., current context) changes only
from segment to segment.

The linguistic, intentional and attentional components are
interrelated.  In particular, the {\it attentional state} describes
the processing of the {\it discourse segment} which has been
introduced to satisfy the current {\it discourse intention}.  The
functional interrelation is expressed temporally in spoken discourse
--- the linguistic, intentional and attentional components devoted to
one DSP co-occur.  Therefore, a change in one component reflects or
induces changes in the rest.  For example, changes ascribed to the
attentional state indicate changes in the intentional structure, and
moreover, are recognized via qualitative changes in the linguistic
structure.  It is because of their interdependence and synchrony that
I can postulate the hypothesis that co-occurring linguistic and
attentional phenomena in spoken discourse --- cue phrases, pauses and
discourse structure and processing --- are linked.

The part of the theory most directly relevant to my investigation are
those constructs that model the attentional state. These are the {\it
focus space} and the {\it focus space stack}.  The {\it focus space}
is the computational representation of processing in the current
context, that is, for the discourse segment currently under
construction.  Within a focus space dwell representations of the
entities evoked during the construction of the segment ---
propositions, relations, objects in the world and the DSP of the
current discourse segment.

A focus space lives on a pushdown stack called the {\it focus space
stack}.  The progression of focus in a discourse is modeled via the
basic stack operations --- pushes and pops --- applied to the stack
elements.  For example, {\it closure} of a discourse segment is
modeled by {\it popping} its associated focus space from the stack;
{\it introduction} of a segment is modeled by {\it pushing} its
associated focus space onto the stack; {\it retention} of the current
discourse segment is modeled by leaving its focus space on the stack
in order to add or modify its elements.

The contents of a focus space whose DSP is satisfied are accrued in
the longer lasting intentional structure.  Thus, at the end of a
discourse the focus space stack is empty while the intentional
structure is fully constructed.

The focus space model abstracts the processing that all participants
must do in order to accurately track and affect the flow of discourse.
Thus, it treats the emerging discourse structure and the changing
attentional foci as publicly accessible properties of the discourse.
However, although the participants themselves may act as if they are
manipulating public structures, the informational and attentional
properties of a discourse are, in fact, modeled only privately.

In explaining certain lexical and prosodic features of discourse, it
is often useful to return to these private models.  For a speaker's
utterance is conditioned both by the state of the her own model and by
her beliefs about those of her interlocutors.  The time--dependent
nature of speech emphasizes the importance of synchronizing private
models.  Lexical and prosodic focusing cues hasten synchronization. In
particular, they guide the listeners in updating their models (among
them, the focus space stack) to reflect the attentional changes
already in effect for the speaker.

For my analysis, the most relevant private model belongs to the {\it
current speaker}, whose discourse intentions guide, for the moment,
the flow of topic and attention in a discourse and whose spoken
contributions provide the richest evidence of attentional state.  If
cue phrases and unfilled pause durations can be shown to correlate
with attentional state (and by definition, the intentional and
linguistic structure), the attentional state they reveal belongs to
the current speaker, and the attentional changes they denote are the
ones the speaker makes in her own private model.

\section{Main Hypotheses}

The theory of the tripartite nature of discourse frames my hypotheses
about the correlation of cue phrases, pause duration and discourse
structure.  The main hypotheses are these: that particular unfilled
pause durations tend to correlate with particular cue phrases and that
this correlation is occasioned by changes to the attentional state of
the discourse participants, or, equivalently, by the emerging
intentional structure of a discourse.

\subsubsection{Cue phrases}
Changes to the attentional state occur at segment boundaries.  Cue
phrases by definition evince these changes --- they are utterance--
and segment-initial words or phrases and they inform on structural or
rhetorical relations rather than on topic.  Thus, for cue phrases, the
question is not whether they correlate with attentional state, but
how.  To answer this question, we ask, for each cue phrase (e.g., {\it
Now, To begin with, So}), whether it signals particular and distinct
changes to the attentional state.

\subsubsection{Pauses}
The correlation of unfilled pauses with attentional state is less
certain because pauses appear at all levels of discourse structure.
They are found within and between the smallest grammatical phrase, the
sentence, the utterance, the speaking turn and the discourse segment.
Their correlation is mainly with the cognitive difficulty of producing
a phrase or utterance\cite{GOLDMAN-EISLER-1}.  To link this
correlation with the task of producing discourse structure, we must
posit a variety of attentional operations with corresponding
variability in cognitive difficulty.  Specifically, we construct the
chain of assumptions that:
\jjitem
{More than one attentional operation exists (e.g., initiation,
retention, closure).}
\jjitem
{The different attentional operations are distinguished by their
effect on the attentional state and by the cognitive difficulty of
their production.}
\jjitem
{The amount of silence preceding an attentional operation is
correlated with the greater or lesser demands it makes on mental
processing.}

To link unfilled pause duration to discourse structure we must first
establish that operations on the attentional state can be
distinguished sufficiently to explain the different demands that each
operation makes on discourse processing and which, therefore, might be
reflected in the duration of segment-initial unfilled pauses.

\section{Auxiliary hypotheses}

The linking of pause duration to the processing of discourse segments
motivates some auxiliary hypotheses that refine notions about the
kinds of mental operations sanctioned by the focus space model and
about the internal structure of a discourse segment.  These auxiliary
hypotheses are developed in this section.

\subsection{Attentional operations}

In the theory of discourse structure, changes to the attentional state
are modeled as operations on the focus space stack.  These operations
appear reducible to four distinct sequences of stack operations that
correspond to four distinct effects on the attentional state, as
follows:
\jjitem
{One push --- {\it Initiate} a  new focus space.}\vsp
\jjitem
{No push, no pop --- {\it Retain} the current focus space.}\vsp
\jjitem
{One or more pops --- {\it Return} to a previously initiated focus
space.}\footnote{When at least one focus space remains on the stack,
the discourse continues.  When none remain, however, the discourse is
ended.}
\jjitem
{One or more pops followed by a push --- {\it Replace} a previous focus
space(s) with a new one.}

The arrangement is asymmetrical in that it is possible to pop more
than one focus space per operation, but to push only one, as shown in
Table~\ref{FocusSpaceOps}.

\begin{footnotesize}
\begin{table}[htbp]
\begin{center}
\begin{tabular}{lccl}
          & Focus space &  Focus space &  \\
          & stack before &  stack after &  \\
Operation &  operation&  operation & Summary\\
\hline
&&&\\
{\bf Initiate} &\fsse{FS}{1} & \fsse{FS}{2}&  One push.\\ 
               &             & \fsse{FS}{1}&           \\
&&&\\
{\bf Retain}& \fsse{FS}{1}&\fsse{FS}{1}&No push, \\
            &             &            & no pop.\\
&&&\\
{\bf Return}  & \fsse{FS}{2} & \fsse{FS}{1}& One or more\\
              & \fsse{FS}{1} &             & pops. \\ 
&&&\\
{\bf Replace} &\fsse{FS}{1}&\fsse{FS}{2} & One or more \\
              &            &             & pops, followed \\ 
              &            &             & by a push. 
\end{tabular}
\end{center}
\caption
{\bf The effect of the four focusing operation on the focus spaces
(FS) in the pushdown focus space stack.}
\label{FocusSpaceOps}
\end{table}
\end{footnotesize}

The decomposition of focus space operations into stack operation
primitives is not merely an attempt to impose a computational patina
on descriptive terms. Rather, it suggests that operations that differ
in kind and number place different requirements on mental processing
for both speaker and therefore might be accompanied by lexical and
acoustical phenomena that also differ.

\subsection{Structure of a discourse segment}

To further motivate the particular usefulness of cue phrases and
unfilled pauses as locators of discourse segment boundaries and
markers of attentional state, it is useful to distinguish among three
phases in the life of a discourse segment (and its focus space
counterpart) --- its initiation, development and closure .  We make
the additional assumptions that each phase may be marked {\it
explicitly} or {\it implicitly} and by {\it lexical} and {\it
acoustical} phenomena.\footnote{Gestural correlates of discourse
structure and processing are outside the scope of this investigation.}

From inspection of dialogue, it appears that the development phase
must be instantiated explicitly with lexical contributions, while the
boundary phases need not be.  However, while lexical marking of
segment boundaries is optional, prosodic marking is not.  Thus, at
initiation of a discourse segment we find, for example, an expanded
pitch range\cite{COOPER-SORENSEN} and at its closure, phrase-final
lowering\cite{HIRSCHBERG-PIERREHUMBERT} and syllable lengthening
\cite{KLATT75}. 

Sometimes, the same structural cue is implicit for one segment yet
explicit for another. For example, in a {\it Replace} operation,
explicitly marked closure of one segment implicitly permits the
initiation of the next.  Conversely, an explicitly marked {\it
Initiation} of the current segment testifies implicitly to the closure
of the previous one.

Boundary phenomena are of special relevance toward retrieving
discourse structure from a multiplicity of lexical and acoustic clues.
The distinction between explicit and implicit correlates for each
phase of segment construction admits four classes of segment boundary
phenomena --- phenomena that are: explicit and segment-initial;
implicit and segment-initial; explicit and segment-final; and implicit
and segment-final.  An investigation of how cue phrases and unfilled
pauses reflect discourse structure and the state of its processing is
thus an investigation of the {\it explicit} and {\it segment-initial}
evidence of focus space initiation.

The selection of segment-initial phenomena in no way implies that
segment-final phenomena are any less crucial to the communication and
recognition of discourse segment boundaries.  Nor does the selection
of cue phrases and unfilled pauses minimize the contributions of other
lexical and prosodic phenomena.  Rather, these selections are
motivated by features of the focus space model that both cue phrases
and unfilled pauses might specially illuminate, and conversely, by
features of the model that might specially illuminate the discourse
function of cue phrases and unfilled pauses. These features are
described in the following two sections.

\section{Cue phrases, discourse markers and attentional state}
\label{CuesAndMarkers}
Cue phrases are those words or phrases which introduce an utterance
--- e.g., {\it To begin with, First of all, Now, But} --- and
coordinate the flow of conversation and focus rather than contribute
directly to the topic at hand.  They provide broad, topic independent
indications of how the speaker intends to relate the current utterance
to those preceding it, thus locating the utterance in the discourse
structure.  The information they convey is attentional, intentional or
both.

The study of cue phrases and their correlation with discourse
structure and focus of attention is most extensive for the {\it
discourse marker}\cite{SCHIFFRIN} subcategory.  Schiffrin's work in
particular, is the basis for my predictions about the structural
effects of cue phrases on the focus space model.

\subsection{Discourse markers}
Discourse markers are generally single word phrases, such as {\it Well,
Now, Then} or {\it So}, whose pragmatic role in a discourse usually
follows from their syntactic and semantic role in a grammatical phrase.
That is, if a word in semantic guise relates {\it propositions in a
grammatical phrase}, it marks in its pragmatic guise the same or similar
relation between {\it utterances in a discourse}.  For
example\cite{SCHIFFRIN}:
\jjitem 
{{\bf And}, as a discourse marker, indicates connectedness, conveying
the speaker's view that the utterance it heads is connected to the
prior discourse.  The connection may be to the immediately previous
utterance or to the speaker's prior [interrupted] turn.}
\jjitem 
{{\bf But} also a marks connectedness, but connects utterances in a {\it
contrast} relation. The contrast may be structural (resumption after a
digression or interruption) or rhetorical.  Like {\bf well}, it
introduces unexpected or undesired material, but in a less cooperative
manner.}
\jjitem
{{\bf I mean} precedes a repair or modification of the speaker's own
contribution or highlights something to which the speaker believes the
hearer should attend.}
\jjitem
{{\bf So} may precede a presentation of a result, and indicates
transitions to a higher level, in contrast to ``{\bf because}'' which
indicates progressive embedding.}
\jjitem 
{{\bf Now} emphasizes what the speaker is about to do, and is often
used to introduce evaluations.}
\jjitem
{{\bf Well} is often used in response, when the possibilities offered
by the previous speaker are inadequate.  It indicates an awareness of
conversational expectations but also heralds a violation of the
previous speaker's expectations.}
\jjitem 
{{\bf You know} indicates an appeal to shared knowledge and mutual
beliefs.}

\subsection{Discourse markers reinterpreted}
\label{structural-correspondences}
Some of the observations about the conversational role of discourse
markers invoke structural effects (embedding, return to a higher
level) although without detailing the structure in question.  A more
unified and computationally driven account might be posed in terms of
operations on the focus space stack, as follows:
\jjitem 
{{\bf And} (connectedness): {\it Retain}, {\it Return}.}
\jjitem 
{{\bf But} (contrast): {\it Retain}, {\it Replace} or {\it Return}.}
\jjitem
{{\bf I mean} (modification or repair): {\it Initiate}, {\it Retain}.}
\jjitem 
{{\bf So} (presentation of a result): {\it Return}, {\it Replace}.}
\jjitem
{{\bf Because} (progressive embedding): {\it Initiate}.}
\jjitem {{\bf Now} (what the speaker is about to do): {\it Replace}.}
\vsp 
\newline
\jitem 
{{\bf Well} (inadequate options):  {\it Replace}.}
\vsp 
\newline
\jitem 
{{\bf You know} (appeal to shared knowledge): {\it Retain}, or {\it
Initiate} when it precedes an aside.}

In addition, there are the cue phrases that highlight structural or
propositional ordinality.  The first use of such a phrase (e.g.,
{\it To begin with, In the first place,}) is likely to denote a
focus space {\it Initiation} while subsequent uses (e.g., {\it
Secondly, Finally,}) denote a focus space {\it Replacement}.

These formulations are not deterministic.  They illustrate, however,
the hypothesis that certain of the discourse markers are more likely
to betoken certain focusing operations.  Under what conditions might
such correspondences exist?  Clearly, features of the context in which
a cue phrase is used might constrain its effect on focusing, and so
explain how conversants are able to track focus from cues that, by
themselves, are ambiguous.

Thus, to select the probable from the possible, corroboration from
other quarters is required. Lexical corroboration may be semantic,
from domain specific evidence of topic change or continuation.  Or it
may be syntactic, from those syntactic distributions that tend not to
cross segment boundaries (tense, aspect and the scope of referring
expressions\cite{GROSZ-SIDNER})\footnote{For example, Walker and
Whittaker observe that deictic pronominal reference may cross segment
boundaries, while nondeictic pronominal reference does so only
rarely\cite{WALKER-WHITTAKER}.} Alternatively, prosodic features are
likely to better identify the current use of a cue phrase from those
that are possible.

\section{Unfilled pauses and attentional state}

The most useful prosodic correlates of discourse segmentation occur
at segment boundaries and indicate either the opening of a new
segment, closure of the old or both.  For example, a phrase-final
continuation rise forestalls segment closure while phrase-final
lowering confirms it\cite{PIERREHUMBERT-HIRSCHBERG}.  And expanded
pitch range tends to mark the introduction of new topics, while
reduced pitch range marks subtopics and parentheticals.  Similarly,
voice quality changes, e.g., from normal to creaky voice, may
accompany attentional and intentional changes.

Filled pauses (e.g., {\it Um, uh}) and unfilled pauses appear at
segment boundaries but are also found within a discourse segment and
in the smaller groupings it contains.  In contrast to the propositional
and attentional accounts of intonational
cues\cite{PIERREHUMBERT-HIRSCHBERG}, accounts of pausing invoke the
demands of cognition and pragmatics. For example, the duration of
unfilled pauses has been observed to correlate with the cognitive
difficulty involved in producing an
utterance\cite{GOLDMAN-EISLER-1}, while filled pauses may function
as a floor holding device\cite{MACLAY:OSGOOD}, or perhaps, correlate
with the speaker's emotional response to
topic\cite{GOLDMAN-EISLER-1}.

As corroborators of attentional interpretations of cue phrases filled
pauses are less useful than unfilled pauses because they overlap with
cue phrases in both form (partially lexicalized) and function. A more
independent measure is provided by unfilled pauses which are not
lexicalized and therefore carry neither lexical nor intonational
propositions.  Rather, as correlates of the cognitive processing, they
may also correlate with the specific differences among stack
operations, which, after all, are cognitive operations, albeit
idealized.

The selection of unfilled pause duration as a possible marker of
attention and segmentation also has the practical advantage of being
easy to locate instrumentally and easy to check perceptually.
Moreover, its measurement is unambiguous instrumentally and requires
less from perception, than, for example, intonational prosodic cues.
For, while intonational features are categorical according to their
type (combinations of the L, H and * tokens\cite{PIERREHUMBERT}) and
the structure to which they apply (word, intermediate phrase,
intonational phrase), pause duration is ordinal and is measured on the
same continuous linear scale for all levels of linguistic and
intonational structures.

\section{Questions and predictions}
My investigation is inspired by the theory relating attentions,
intentions and discourse structure\cite{GROSZ-SIDNER}.  To the more
specific observations linking cue phrases to attentional
state\cite{GROSZ-SIDNER,SCHIFFRIN} and the duration of unfilled pauses
to increased cognitive difficulty\cite{GOLDMAN-EISLER-1}, I add the
assumption of four fundamental focusing operations.  Together, they
motivate my hypotheses that:

\jenum{(1)}{Specific cue phrases betoken specific focusing operations.}
\vsp\newline
\jenum{(2)}{Differences in the cognitive difficulty of the focusing operations
are reflected in the duration of the pauses that precede them.}
\newpage
From these hypotheses come the specific questions that guide the
research:
\jjitem
{Is there a correlation between the focusing operations and the duration
of the pause that precedes it?}
\jjitem
{Are cue phrases correlated with focusing operations --- how often and
under what circumstances?}
\jjitem 
{What is the relation of pausing and cue phrases --- do they
substitute for each other, compliment each other or play
different roles such that one is required or allowed where
the other is not?}
\jjitem 
{Is there a unique minimum cognitive cost for each stack primitive
(Push, Pop) of which focusing operations are composed, and that would
therefore explain differences in segment-initial pause duration?}
\vsp 

In addition, the hypotheses raise questions not immediately
answerable:
\jjitem
{If there are indeed patterns of usage, do they differ predictably for
different discourse features, for example, by format (monologue or
dialogue) or according to the planning effort (prepared or
extemporaneous) required in formulating each utterance?}
\jjitem
{If on the other hand, correlations are partial at best, can other
lexical or prosodic features provide the missing correlates?}

Research into these questions is not without its biases.  Thus, I
expected to find in my discourse samples the following correlations:
\jjitem
{\em Unfilled pause duration and focusing operation are correlated.}
\jjitem
{{\em Cue phrases are correlated with focusing operations.} (The
particular predictions are discussed previously in Section 5.2.)}
\jjitem
{\em Cue phrase type and unfilled pause duration are correlated as well.}

The hypothesized correlation of unfilled pause duration with focusing
operations is based on assumptions about variations in complexity
among the operations, such that longer pauses will accompany more
complex operations.  Complexity is conjectured to correlate with kind
and number.  That is, it varies according to whether the operation
decomposes into pops, a push or both and it increase with the number
of segments opened or closed in one operation.  

This produces the particular predictions that: 
\jjitem
{{\em Retentions will be preceded by pauses of the smallest
duration} because they induce neither a push nor pop and therefore are
the least costly of the focusing operations.}
\jitem
{{\em Pause duration is positively correlated with the number of
segments affected in one focusing operation.} That is, the more
segments opened or closed, the longer the preceding pause.}
\jjitem
{{\em Pops are more costly than pushes.} This follows from an
assumption that adding information (a push) builds on what is
currently established and accessible, while removing information (one
or more pops) makes the production of subsequent utterances more
difficult.}

\section{Data}

I analyzed two discourse samples --- three minutes of a directions
discourse and seven minutes of a manager--employee project meeting.  The
segmentation of the second proved difficult and is still in progress, so I
report results only for the first.

In the directions discourse, Speaker B provides Speaker A with walking
directions to a location on the M.I.T. campus.  The discourse takes
the form of an expert-client dialogue.  Although Speaker A initiates
the dialogue, most of the discourse segments and their intentions are
introduced by Speaker B, the expert.\footnote{The conversation
occurred in a face-to-face encounter and was recorded on a hand-held
cassette recorder.}

\section{Methods}
The search for correlations among cue phrases, unfilled pauses and
discourse structure generated three data collection tasks:
\jjitem
{ Identification of cue phrases;} 
\jjitem
{ Identification and measurement of unfilled pauses;} 
\jjitem
{ Segmentation of the discourse via the identification of the focusing
operations that effected the segmentation.}

\subsection{Cue phrase identification}
The main challenge of cue phrase identification lay in distinguishing
cue from non-cue uses of a phrase.  Usually, cue uses are utterance-
or segment-initial, while non-cue uses are not.  However, this is not
a reliable criterion for the connectives, {\it And} and {\it But},
which may head an utterance or phrase as either a cue phrase or a
syntactic conjunctive. In cases where the usage was unclear, I decided
against the pragmatic usage if the phrase in question provided
syntactic coordination of two semantically related propositions.  If
even this judgment proved difficult, I applied the intonational
criteria that distinguished cue and non-cue uses of {\it
Now}\cite{HIRSCHBERG-LITMAN}.  Thus, if the cue phrase candidate was
deaccented, or accented with L* tones or uttered as a complete
intonational phrase, it was classified as a cue phrase.

\subsection{Pause location and measurement}
Pauses were identified by ear and corroborated and measured using the
waveform and the energy track displays of two signal processing
programs.\footnote{ {\it SPIRE}, written for the LISP machine by Victor
Zue's group at M.I.T. and {\it dspB} (digital signal processing workBench)
written for the DECstation by Dan Ellis at the M.I.T. Media Laboratory.}
The locations of all unfilled pauses were recorded, as were their
durations, rounded to the nearest one tenth of a second.

In general, the procedure was straightforward.  The only confusion was
presented by the silence between the closure and release phase of
plosives.  This silence was not counted as a genuine unfilled pause.

\subsection{Discourse segmentation}

An accurate discourse segmentation falls out of an accurate
classification of the focusing operations by which the segments have been
constructed.  The tasks are interrelated and both are difficult.
Therefore, in this section I will discuss in detail the task, its
difficulties and the classification criteria I developed to enhance
the accuracy of my judgments.

\subsubsection{The task}
The segmentation of a completed discourse is equivalently the task of
recapturing the attentional state that accompanied each successive
utterance.  Attentional cues are especially important because topical
relations do not always predict discourse structure. The points at
which discourse structure diverges from the organization of
information in the domain may be precisely the points at which
attentional cues are most appropriate.

Segmentation of a completed discourse is most straightforward for
expository text. In such discourse, domain and attentional hierarchies
often coincide --- the relations among segments and of each segment to
the overall Discourse Purpose are clear.  In spoken and impromptu
discourse, however, the alignment of DSPs is not always so felicitous.
Even in the task-oriented directions discourse, the relations among
steps in the task did not conclusively determine the relations of the
discourse segments in which these steps were described.

The particular segmentation difficulties presented by my sample(s) led
to the development of explicit criteria for isolating the
corroborating features of attentional operations and discourse
structure.  The criteria help clarify confusion from two sources ---
the distinction between attentional and domain hierarchies and the
interpretation of underspecified lexical and prosodic attentional
cues.

\subsubsection{Separating the attentional from the topical.}

In prepared discourse (written or spoken) the intentional structure
is tightly coupled to the Discourse Purpose.  In contrast, impromptu
discourse exhibits a looser coupling, owing to its real-time and
situated nature.  In such discourse, the maintenance of coherence
requires the real-time management of cognitive resources upon which
competing demands may be made.  As a consequence, influences outside
the ostensible DP must be managed in support of continuing the
conversation at all.  Because DSPs that are ostensibly outside the
current DP can become temporarily relevant, provision must be made
for their principled incorporation into the attentional state and in
the linguistic and intentional structures.

This is accomplished via attentional constructions that are more
likely to occur in spoken discourse, for example, flashbacks,
digressions and interruptions\cite{GROSZ-SIDNER}.  Their relation to
the discourse in which they occur illustrates the difficulty of
segmenting in hindsight a discourse whose DSPs may satisfy multiple
DPs.  This recommends against reliance on domain knowledge, since one
discourse may invoke more than one domain.

Therefore, to locate segment boundaries, I use criteria that
emphasize focusing operations independent of the ostensible DP.  For
example, although the succession of two topically unrelated segments
might suggest a {\it Replace} operation, it is treated as an {\it
Initiate} in the presence of explicit indicators of linkage or in the
absence of explicit indicators of separation.  Consequently, 
successive segments may be linked hierarchically in the attentional
and linguistic structures despite their topical independence.

For example, in the following section of the directions
discourse {\tt (1)} is a topic introduction, {\tt (2)} a digression
and {\tt (3)} an elaboration, i.e., a subtopic:

\begin{small}
\hspace*{.2in}{\tt (1)} \parbox[t]{2.5in} 
{To your left,}
\newline
\hspace*{.2in}{\tt (2)} \parbox[t]{2.5in} 
{if you have followed these directions faithfully,}
\newline
\hspace*{.2in}{\tt (3)} \parbox[t]{2.5in} 
{y'know you'll be facing a wall straight ahead of you,}
\end{small}

Although {\tt (2)} is a comment on discourse processing, it functions
neither as a cue phrase nor a synchronization device.  The digression
it represents it not topically subordinate to {\tt (1)}, nor is {\tt
(3)} topically subordinate to {\tt (2)}.  However, they are
attentionally subordinate to the utterances they follow, as indicated
by the continuation rises at the end of {\tt (1)} and {\tt (2)}.
While the semantic and topical differences between successive
utterances argue for segment separation, the acoustical concomitants
argue against.  Therefore, the attentional moves that introduce {\tt
(2)} and {\tt (3)} contain no pops. Instead, they are {\it
Initiations}, producing the following segmentation:

\begin{small}
\parbox[t]{.35in}{\it Replace}\hspace*{.1in}{\tt (1)}
\parbox[t]{2.3in} {To your left,}
\newline
\parbox[t]{.35in}{\it Initiate}\hspace*{.1in}{\tt (2)}\hspace*{.2in}
\parbox[t]{2.1in} {if you have followed these directions faithfully,}
\newline
\parbox[t]{.35in}{\it Initiate}\hspace*{.1in}{\tt (3)}\hspace*{.4in}
\parbox[t]{1.9in} {y'know you'll be facing a wall straight ahead of you.}
\end{small}

\subsubsection{Interpreting underspecified cues}

Even when successive utterances are aligned attentionally and
topically, their cue phrase and prosodic markings may not
conclusively reveal their exact place in discourse structures.  The
underspecified nature of cue phrase correspondences to focusing
operations is discussed in Section 5.2.  Prosodic marking is
similarly underspecified, and on two counts.  First, a particular
intonational feature at the (e.g., phrase-final lowering,
phrase-initial pitch range expansion) can felicitously indicate more
than one focusing operation; second, the intonation at a phrase
boundary often indicates stack primitives (push, pop, null) more
reliably than the composite focusing operations from which discourse
structure is deduced.

For example, in the directions discourse, the cue phrases {\it So,
But} and {\it And} often indicated pops, as did the prosodic changes
that accompanied them, e.g., expanded pitch range and a shift from
L* to H* tones.  However, these cues did not reveal exactly how many
segments were popped nor whether a push followed the sequence of
pops.  Thus, it was not always easy to distinguish a {\it Return}
(one or more pops) from a {\it Replace} (one or more pops, followed
by a push).

Neither domain nor syntactic knowledge were conclusive in this
regard.  For example, domain and syntax dictated the following
segmentation:

\begin{small}
\parbox[t]{.35in}{\it Return} \hspace*{.1in}{\tt (4)}
\parbox[t]{2.3in} 
{And you need to turn left and then walk along Building Five.}
\newline
\parbox[t]{.35in}{\it Initiate} \hspace*{.1in}{\tt (5)}\hspace*{.2in}
\parbox[t]{2.3in} 
{And you'll be  walking through the architecture lofts.}
\end{small}

but in contraindication to what was specified intonationally:

\begin{small}
\parbox[t]{.35in}{\it Return}\hspace*{.1in}{\tt (4)}
\parbox[t]{2.3in} 
{And you need to turn left and then walk along Building Five.}
\newline
\parbox[t]{.35in}{\it Retain}\hspace*{.1in}{\tt (5)}
\parbox[t]{2.3in} 
{And you'll be  walking through the architecture lofts.}
\end{small}

(The intonationally driven segmentation, in contradiction to the
structure of knowledge in the domain, may account for the listener's
subsequent confusion about the very point made in this section of the
discourse.)

\subsubsection{Classification criteria}

Because semantic clues to attentional state can be confusing and
lexical and prosodic markings inconclusive, it became necessary to
standardize the procedure and criteria for classifying the focusing
operations.  An accurate classification depends on the answers to two
questions for the phrase undergoing classification: Has a new focus
space been opened?  Has an old focus space been closed?  Most useful
in this regard are the lexical and prosodic phenomena within and
around the phenomena currently under evaluation for their attentional
effect.

What constitutes current phenomena, and what might constitute its
surrounds?  I selected as {\it current} the speech fragment that
begins with one of five fragment-initial tokens and whose terminating
boundary is marked by the occurrence of the next fragment-initial
token.  These tokens are:
\newline\jitem{The unfilled pause;}
\newline\jitem{The filled pause;}
\newline\jitem {A cue phrase;}
\newline\jitem{An acknowledgment form: {\it Ok, Sure, Uh-huh,} etc.;}
\newline\jitem 
{Or the unmarked case: any other sentence-initial grammatical
constituent, e.g., a noun phrase, auxiliary verb, complementizer or
adverb.}

My demarcation of the relevant surrounding phenomena was less bound to
structure than to function.  For both prior and subsequent phenomena,
I selected the smallest speech fragment that could be distinguished by
its discourse function, i.e., by its attentional, coordination or
topical role.  I assign five classifications:
\newline\jitem {A cue phrase;}
\newline\jitem{An acknowledgment or prompt;} 
\newline\jitem{A segment closure (e.g., {\it Good!});}
\newline\jitem {A repair;}
\newline\jitem {Or the unmarked case --- development of the topic.}

The lexical and acoustical features of prior, current and subsequent
speech fragments are treated as corrobating evidence for the
attentional operation associated with the {\it current speech
fragment}.  Often this evidence indicated a stack primitive --- push
or pop --- rather than a full-fledged focusing operation.  This is
illustrated in Table~\ref{Features}, which catalogues the lexical and
prosodic features exhibited by prior, current and subsequent speech,
and the stack and focusing operations for which each is considered
evidence.
\begin{footnotesize}
\begin{table}[htbp]
\begin{center}
\begin{tabular}{l   l  l l}
\ul{\bf GIVEN:}&        &\ul{\bf CONCLUDE:}        & \\
         &        &        &{\sc Focusing}       \\
         &        &        & {\sc Operation}      \\
{\sc Speech}&        &{\sc Stack}  &  {\sc for Current}  \\
{\sc Evidence} &{\sc Feature}&{\sc Primitive} & {\sc Fragment}        \\
\hline\hline
{\bf Prior}& Falling phrase-final   & Pop of       & Replace. \\
{\bf speech}& intonation,           & co-occurring& \\
&acknowledgment, & segment(s).  & \\
&lexical/semantic closure.&  & \\
&& & \\
& Phrase-final      &  Null &  Retain.\\
& continuation rise.&       & \\
&& & \\
\hline
{\bf Current}& Pronominalization,     & Push             &  Initiate, \\
{\bf speech}& reduced pitch range,    & or Null          &  Retain. \\
& nonstandard phonation,              & for              &\\
& many L* accents                     & co-occurring    &\\
& (parentheticals),                   & segment.         & \\
& relative clause,                    &                  & \\
& {\it Now, Y'know,}                  &                  &\\
&  Ordinal cue phase.                 &                  &\\
&& &\\
& Nonpronominalized        & Pop of    &  Return,\\
& repetition (e.g., segue),& previous  &  Replace.\\
& expanded pitch range,    & segment(s).& \\
& reintroduction of        & & \\
& normal phonation,        & & \\
& {\it So, But}.           &  &\\
&& & \\
& Falling phrase-final    & Impending     &  Retain\\
& intonation,            & Pop of        & (but an \\
& acknowledgment,        & co-occurring  & impending\\
& prompt,                & segment(s).   & Return \\
& lexical closure,       &               & or \\
& phrase-final creaky    &               &Replace).      \\
& voice.                 &               & \\
&& & \\
\hline
{\bf Subsequent}& Nonpronominalized            & Pop of      &  Return,\\
{\bf speech}    & repetition (e.g, a segue),   & previous    &  Replace.\\
& expanded pitch range,        & segment(s). & \\
& normal phonation,            & & \\
& {\it So, But, Now}.          & & \\
\end{tabular}
\end{center}
\caption
{{\bf The lexical and acoustical features that support
classifications of stack primitives and focusing operation(s).} A
co-occurring segment denotes the segment containing the speech
(prior, current, subsequent) under examination. The focusing operations,
however, describe the attentional interpretation that such speech
indicates for the {\it current} speech fragment.}
\label{Features}
\end{table}
\end{footnotesize}

\label{fragments}

\subsubsection{Coding the data}
The data relevant to every speech fragment was coded for later
statistical analysis.  This translated into two tasks --- identifying
the prior, current and subsequent speech fragment and for each current
fragment, recording:
\jjitem
{The duration of the preceding unfilled pause;}
\jjitem
{The type of fragment-initial constituent, either:}
\newline
\kitem{A cue phrase;}
\newline
\kitem{An explicit acknowledgment form (e.g., {\it Ok, Sure.});}
\vsp\newline
\kitem{A filled pause;}
\newline
\kitem{Or any other sentence-initial syntactic form whose function
is primarily topical, not pragmatic.} 

\jitem{The co-occurring focusing operation.}

\jitem
{The embedding of the current segment in the linguistic structure
(number of levels).}

\jitem
{The number of segments opened or closed in the focusing operation.}

\jitem
{The discourse function of the immediately prior speech (cue phrase,
acknowledgment, closure, filled pause, repair, topical but none of
the above).}
\newpage
\jitem
{The discourse function of the immediately subsequent speech (same
categories as for prior speech).}
\jjitem
{Whether the speaker was initiating or continuing a speaking turn
with the current fragment.}

Using this metric, one hundred speech fragments were identified
according and their features coded.  The coded representation of the
discourse was then analyzed for distributions and statistical
correlations.\footnote{The discourse function classifications and the
within-/between-turn distinctions were recorded to track the features
influencing the judgment of focusing operation, but were not included
in any calculations.} The results are reported in the next section.

\section{Results}
In this section I summarize the raw data, report the results of
statistical tests and offer an explanation of the findings.

\subsection{Data}
The segmentation of the discourse was reconstructed according to the
focusing operations indicated both lexically and acoustically.  The
segmentation described a discourse with two top level segments.
Within the first, the overall task was defined; within the second, it
was executed.  The task definition segment was itself composed of two
top level segments, while the execution segment is composed of nine.

The key elements of the coding scheme were, of course, the focusing
operation, the fragment-initial token and the duration of the unfilled
pause preceding the fragment.  Distributions for these categories are
catalogued in Table~\ref{RawData}.
\begin{footnotesize}
\begin{table}[htbp]
\begin{center}
\begin{tabular}{l|lcc}
{\sc Category}&{\sc Feature}&\multicolumn{2}{c}{\sc Number of Occurrences} \\
\hline\hline
          &                 & {\em Marked} & {\em Unmarked}\\
{\bf Focusing}  & Initiate & 13          & 10\\ 
{\bf operation} & Retain   & 18          & 37\\
                & Return   &{\one}6      &{\one} 5 \\
                & Replace  &\ul{{\one}7} & \ul{{\one}4}\\
           & {\em ALL}     & 44  & 56 \\
&&&\\
\hline
           &                 & {\em Initial} & {\em Internal}\\
{\bf Fragment-}  & And       &{\one}3  &{\one}4 \\
{\bf initial}    & But       &{\one}2  &{\one}1 \\ 
{\bf constituent}& Now       &{\one}2  &{\one}-- \\
           &  Oh             &{\one}2  &{\one}--\\
           &  So             &{\one}3  &{\one}2  \\
           &  Well           &{\one}2  &{\one}--\\
           &  Y'know         &{\one}2  &{\one}--\\
           &Ordinal cue phrase &{\one}1&{\one}--\\
           & Acknowledgment  &{\one}2  &{\one}7\\
           & Filled Pause    &{\one}7  &{\one}4 \\
           & Unmarked  & \ul{19} & \ul{37}  \\
           & {\em ALL}                          & 45  & 54 \\
&&&\\
\hline
          &      {\em Seconds} & {\em Initial} & {\em Internal} \\
{\bf Unfilled}&0.0 &{\one}5 & 15 \\
{\bf pauses}  &0.1&{\one}6  & 11  \\
             &0.2 &{\one}3  & 15  \\
             &0.3 &{\one}4  &{\one}5  \\
             &0.4 &11       &{\one}5 \\
             &0.5 &{\one}1  &{\one}5  \\
             &0.6 &{\one}3  &{\one}4 \\
             &0.7 &{\one}4  &{\one}2  \\
             &0.8 &{\one}1  &{\one}--\\
             &0.9 &{\one}1  &{\one}--\\
             &1.7 &{\one}1  &{\one}--\\
             &2.0 &\ul{{\one}1}  &\ul{{\one}--}\\
             &         & 41 & 62 \\
&&&\\
 & {\em Average}     & 0.422 seconds & .224 seconds \\
\end{tabular}
\end{center}
\caption
{{\bf Distributions of fragment-initial constituents, focusing
operations and pause durations.} Separate counts are taken for {\it
segment-initial} and {\it segment-internal} phenomena and for {\it
marked} and {\it unmarked}.  A {\it marked} focusing operation begins
with a cue phrase, an acknowledgment form or a filled pause, while
an {\it Unmarked} operation does not.}
\label{RawData}
\end{table}
\end{footnotesize}

\subsection{Statistical analyses}
The predictions were analyzed via statistical tests on the coded
representation of the discourse.

\subsubsection{Pause duration and focusing operation}
A comparison of the mean pause duration for each focusing operation
showed a significant difference among the operations (F(3,96)=7.31,
{\it p}$<$.001).  The data in Table~\ref{pause-fssop} point to the
{\it Replace} operation as most different from the other three
operations in this regard.\footnote{However, the importance of this
observation is offset by the small sample size and large standard
deviation.}

\begin{footnotesize}
\begin{table}[htbp]
\begin{center}
\begin{tabular}{lccc}
{\sc Focusing}      & {\sc Number}    &{\sc Mean Pause}         &{\sc Standard } \\
{\sc Operation}& {\sc of Tokens} &{\sc Duration (seconds)} &{\sc Deviation } \\
\hline
{\bf Initiate} &   23 &    0.3217           &    0.2173   \\
{\bf Retain} &     55 &    0.2091           &    0.1818 \\   
{\bf Return} &     11 &    0.2545           &    0.2505 \\  
{\bf Replace} &    11 &    0.6500          &    0.6727 \\  
\end{tabular}
\end{center}
\caption{\bf Mean pause durations for each focusing operation.}
\label{pause-fssop}
\end{table}
\end{footnotesize}

\subsubsection
{Pause duration and number of segments affected in a focusing operation}
                                    
Longer pauses were positively correlated with the number of segments
opened or closed during one focusing operation (r = .357, {\it
p}$<$.001).  This finding might partially explain the long pauses that
appear before a {\it Replace}, since a {\it Replace} is the focusing
operation most likely to affect the most focus spaces, By definition,
it requires [almost] everything to be popped from the focusing before
the initiation (push) of a new focus space.

\subsubsection{Pause duration and depth of embedding}
A correlation of pause duration and the depth of embedding in the
linguistic structure (or equivalently, the number of focus spaces
still on the stack) showed no significant effect on pause duration
(F(1,98) = 0.1861, {\it p}$<$.7).

\subsubsection{Pause duration, cue phrase and focusing operation}
The directions dialogue contained too few fragment-initial tokens to
calculate meaningful statistics about their relation to focusing
operations.  Therefore, the best course was to select from the raw
data (see Table~\ref{mean-values}) the patterns that were likely
candidates for further testing.  For example, {\it So} was never
associated with an {\it Initiate} operation and also was preceded by
the smallest mean pause durations (0.13 seconds). A filled pause, with
a similar mean pause duration (0.14 seconds) was primarily associated
with {\it Initiates} and {\it Retains} but never with {\it Replace}.
And, while {\it And} shared the same focusing operations as a filled
pause, its mean value for pause duration was more than twice as large
(0.33 seconds).
\begin{footnotesize}
\begin{table}
\begin{center}
\begin{tabular} {lccccl}
{\sc Initial} & & & & & \\
{\sc Token} &{\sc Initiate}&{\sc Retain}&{\sc Return}&{\sc Replace}&{\sc ALL}\\
\hline
{\bf And}
&0.43{\two} 3 & 0.25{\two} 2 & 0.25{\two} 2 &--  & 0.33{\thr} 7 \\
{\bf But }
&  --          & 0.70{\two} 1&0.00{\two} 1&0.10{\two}  1&0.27{\thr} 3 \\
{\bf Now}
&   --         &   --  &     --  & 0.55{\two} 2 & 0.55{\thr} 2 \\
{\bf Oh}
&  --          & 0.00{\two} 2&   -- &    --  & 0.00{\thr} 2  \\
{\bf So}
&  --          &0.15{\two} 2 & 0.15{\two} 2 & 0.05{\two} 1 & 0.13{\thr} 5 \\
{\bf Well} 
&  --  & -- & -- & 0.20{\two} 2 & 0.20{\thr} 2 \\
{\bf Y'know }
& 0.40{\two} 2  & -- & -- & -- &  0.40{\thr}  2\\
{\bf Ordinal}  
& 0.40{\two} 1&  --  &  --   &  --  &0.40{\thr} 1\\
{\bf Acknow--}
&0.10{\two} 1 & 0.20{\two} 7 & -- & 0.90{\two} 1 & 0.27{\thr} 9 \\
{\bf ledgment}   & & & & & \\
{\bf Filled}
& 0.23{\two} 6 &  0.05{\two} 4 & 0.00{\two} 1 & -- &0.14{\two} 11 \\
{\bf Pause}  & & & & &  \\
{\bf Unmarked} 
&\ul{0.35}{\one} \ul{10}&\ul{0.23}{\one} \ul{37}&\ul{0.40}{\two} \ul{5}
&\ul{1.15}{\two} \ul{4}&\ul{0.33}{\two} \ul{56}\\
{\em ALL}
&0.32{\one} 23&0.21{\one} 55&0.26{\one} 11&0.65{\one} 11&0.29{\one} 100\\
\end{tabular}
\end{center}
\caption
{{\bf The mean duration, in seconds, of the pause preceding
fragment-initial tokens and focusing operations that co-occur.} The
number of tokens in the calculation follows the mean value.}
\label{mean-values}
\end{table}
\end{footnotesize}

\subsubsection{Pause duration and marked/unmarked}

To compensate for the small sample sizes of the cue phrase data, all
explicit lexical markers of structure (cue phrase, acknowledgment,
filled pause) were collapsed into the category, {\it marked}. The data
in this category were compared to the data for lexically {\it
unmarked} fragments.  Because the longest pauses preceded unmarked
{\it Returns} and {\it Replacements}, I predicted that unmarked
operations would in general be preceded by longer pauses than marked.

The results are in the direction predicted and are summarized in
Table~\ref{marked-unmarked}.  The average duration for pauses preceding a
marked focusing operation was 0.24 seconds (standard deviation =
0.24), while the average for pauses  preceding unmarked
operations was 0.33 seconds (standard deviation = 0.36).
Statistically this approaches significance (T(96) = 1.58, {\it p} =
.12).

\subsection{Discussion}

Thus far, analysis of the data identifies significantly longer pauses
for the {\it Replace} operation than for any other and shows that
pause duration is positively correlated with the number of segments
affected by one focusing operation. These findings begin to
distinguish the focusing operations quantitatively, by number of focus
spaces affected, and qualitatively, by whether they occur within an
established context ({\it Initiate, Retain, Return}) or at its
beginning ({\it Replace}).

Although, the raw data in Table~\ref{RawData} appears to show patterns
for specific segment-initial tokens, the number of tokens is
insufficient for establishing a correlation between cue phrase and
focusing operations, let alone a three-way relationship among cue
phrase, pause duration and focusing tasks.

The categorical classification present particular problems.  For,
uncertainties arose even with the application of a classification
metric.  Perhaps these uncertainties should have been incorporated
into the coding scheme or perhaps the categorical classifications
should have been abandoned\footnote{at least, in this stage of the
investigation} in favor of additional and quantifiable acoustical and
lexical features.

\subsection{Refining the original hypotheses} 

Only partial conclusions can be drawn from the data.  However, the
results are useful toward refining the original hypotheses and
determining the content of future research.  The distinction between
the pause data for marked and unmarked fragments is a case in point.
For each focusing operation, the difference between mean pause
durations at best only approaches significance (see
Table~[24~\ref{marked-unmarked}).  However, because the values for all
focusing operations are always greater for unmarked utterances, a
hypothesis is suggested: that, given a speech fragment and the
focusing operation it evinces, the preceding unfilled pause will be
longer if the fragment is lexically unmarked.

\begin{footnotesize}
\begin{table}
\begin{center}
\begin{tabular} {lccccl}
{\sc Speech} & & & & & \\
{\sc Fragment}&{\sc Initiate}&{\sc Retain}&{\sc Return}&{\sc Replace}&{\sc ALL}\\
\hline
{\em Marked}
& 0.30{\one} 13& 0.17{\one} 18&0.13{\two} 6 & 0.36{\two} 7&0.24{\two} 44\\
{\em Unmarked }
&\ul{0.35}{\one} \ul{10}&\ul{0.23}{\one} \ul{37}&\ul{0.40}{\two} \ul{5}
&\ul{1.15}{\two} \ul{4}&\ul{0.33}{\two} \ul{56}\\
{\em ALL}
&0.32{\one} 23&0.21{\one} 55&0.26{\one} 11&0.65{\one} 11&0.29{\one} 100\\
\end{tabular}
\end{center}
\caption
{{\bf The mean duration, in seconds, of the pause preceding focusing
operations and marked or unmarked speech fragments that co-occur.}
The number of tokens in the calculation follows the mean value.}
\label{marked-unmarked}
\end{table}
\end{footnotesize}

If this hypotheses is correct, two accounts can be constructed that
would jointly predict the appearance of cue phrases. One account
emphasizes the processes involved in choosing and communicating the
state of global focus.  The other emphasizes the mutually recognized
(by speaker and hearers) attentional and intentional state of the
discourse.  Together they identify the factors that would impel a
speaker to precede an utterance with a cue phrase, an unfilled pause
or both.

\subsubsection
{The influence of the speaker's internal processes and conversational
goals}

If an unfilled pause preceding a lexically unmarked fragment is
significantly longer, we might assume that a particular focusing
operation is executed in a characteristic amount of time (given
adequate consideration of other contextual features).  Within this
time, we might observe  silence, a cue phrase  or both.
\footnote{The discussion will focus on cue phrases, even
though the points are relevant to other lexical markers of discourse
structure and processing.}

Because both pause and cue phrase can appear at the same location in a
phrase, we ask if their functions are equivalent, or instead,
complementary.  My hypothesis selects the second option, that they are
complementary in the cognitive processing each reflects and in the
discourse functions each fulfills.  For, if the duration of an
unfilled pause is evidence of the difficulty of a cognitive task, a
cue phrase is evidence of its partial resolution.

As a communicative device, cue phrases are more cooperative than
silence. In silence, a listener can only guess at the current contents
of the speaker's models.  With the uttering of a cue phrase, the
listener is at least notified that the speaker is constructing a
response.  The minimal cue in this regard is the filled pause.  {\it
Bone fide} cue phrases, however, herald not only an upcoming
utterance, but a particular direction of focus and even a
propositional relation between prior and upcoming speech.

Cue phrases serve not only the listener but also the speaker.  Because
they commit to topic structure, but not to specific referents and
discourse entities, they buy additional time for the speaker in which
to complete a focusing operation and formulate the remainder of the
utterance.

\subsubsection{The influence of the state of the discourse}
The account of the influence of the currently observable state of the
discourse rests on two patterns in the data: (1) the difference in
pause durations for marked and unmarked {\it Initiate}s and {\it
Retain}s is minimal; and (2) the difference between marked and
unmarked {\it Return}s and {\it Replace}s is greater.  If these
patterns can be shown to be significant, they suggest that remaining
in the current context is less costly than returning to a former
context, or establishing a new one.  The corollary is the claim that
an expected focusing operation need not be marked, while an unexpected
operation is most felicitous when marked.

In other words, remaining in the current context or entering a
subordinate context is expected behavior, while exiting the current
context is not.  Exiting the current context (focus space) carries a
greater risk of disrupting a mutual view of discourse structures.  The
extent of risk is assessed for the listener by the difficulty of
tracking the change and for the speaker, by the difficulty of
executing it.  The risk originates in the nondeterministic definitions
of {\it Return} and {\it Replace} operations --- both contain in their
structure one or more pops.  In addition, these operations can be
confused because both begin identically, with a series of pops.

Because closing a focus space is a marked behavior, the clues to
changing focus are most cooperative if they guide the listener toward
re-invoking a prior context (i.e., a {\it Return}) or establishing a
new one (Replace).  Thus, certain clues are more likely to mark a
return to a former context (e.g., {\it So, Anyway, As I was saying}),
while others ({\it Now}, the ordinal phrases) mark a {\it Replace}.

\subsubsection{Future work} 
The goal of future investigations is to establish the bases for
predicting the appearance of particular acoustical and lexical
features.  The speculations presented in this section provide a
theoretical framework. If borne out, they can be re-fashioned as
characterizations of the circumstances in which cue phrases and
unfilled pauses are most likely to be used.

\section{Conclusion}
The relationships among cue phrases, unfilled pauses and the
structuring of discourse are investigated within the paradigm of the
tripartite model of discourse.  Within this model, the postulation of
four focusing operations provides an operational framework to which
can be tied the discourse functions of cue phrases and the cognitive
activity associated with the production of an utterance.  Especially,
the difficulty of utterance production might be explained by the
complexity of the co-occurring focusing operation.  Such a
correspondence is, in fact, suggested by the positive correlation of
pause duration and the number of focus spaces opened or closed in one
operation on the focus space stack.

However, because the classification of focusing operations is
uncertain, more data and better tests are required to characterize the
relationships among the lexical and acoustical correlates of topic and
focus.  In addition, the aptness of the tripartite model itself is not
assured.  The idealizations it contains may undergo modification in
light of new results, or be augmented by other accounts of discourse
processing.  On the other hand, the analysis of more quantitative data
may confirm the implications of the model, and its appropriateness as
the foundation for investigating the lexical and prosodic features
of discourse.

\section{Acknowledgments}
Many thanks to Susan Brennan who selected and ran the statistical
tests on the data and to S. Lines for helpful comments.
Various stages of this work were supervised in turn by
Chris Schmandt and Ken Haase, both of the M.I.T. Media Laboratory.

\end{document}